\documentclass[5p,authoryear]{elsarticle} 

\usepackage{framed,multirow}
\usepackage{amssymb}
\usepackage{latexsym}
\usepackage{xcolor}
\usepackage{lineno,hyperref}
\modulolinenumbers[5]
\usepackage{amsmath}
\usepackage{multirow}
\usepackage{color,graphicx}
\usepackage{nicematrix}
\usepackage{arydshln}
\usepackage{hyperref}
\usepackage{url}
\usepackage{footnote}
\usepackage{bm}
\usepackage{cite}
\usepackage{subfigure}
\usepackage{verbatim}
\usepackage{ulem}
\usepackage{graphicx}
\usepackage{pgfplots}
\usepackage{cancel}

\usepackage{booktabs}  
\usepackage{tabularx}
\usepackage{booktabs}
\usepackage{graphicx}
\usepackage{amssymb}
\usepackage{makecell}

\usepackage{enumitem}

\journal{Medical Image Analysis}

\begin{document}

\begin{frontmatter}

\title{Learning to Beat: Phenotype-Guided Latent Flow with Regional Motion Priors for Biventricular Motion Synthesis}

\author[label1]{Xuan Yang} 
\author[label1,label2]{Xiaohan Yuan} 
\author[label1]{Hao Li} 
\author[label3]{Lingyu Chen} 
\author[label1]{Yanan Liu} 
\author[label1]{Qingya Li}
\author[label1]{Lei Li*} 
\ead[url]{lei.li@nus.edu.sg}

\address[label1]{Department of Biomedical Engineering, National University of Singapore, Singapore}
\address[label2]{School of Automation, Southeast University, Nanjing, China}
\address[label3]{School of Computer Science and Technology, Nanjing University of Aeronautics and Astronautics, Nanjing, China}

\begin{abstract}

Full-cycle biventricular geometry is essential for characterizing cardiac function. However, dense and temporally consistent 3D+t biventricular meshes are not routinely available, whereas end-diastolic (ED) anatomy can often be obtained reliably. We therefore investigate full-cycle biventricular motion synthesis from a single ED mesh. This task is challenging because cardiac deformation is spatially heterogeneous and phenotype dependent, while conventional global generative models often obscure localized motion patterns. In this study, we propose a region-specific and phenotype-adaptive framework that integrates motion-informed functional parcellation with conditional latent flow. A functional partition learned from reconstructed motion organizes the ventricular surface into regions with coherent dynamics and enables topology-aware regional feature exchange. A phenotype-conditioned rectified-flow model subsequently maps the ED anatomy to full-cycle motion latents through fine-grained conditioning and prototype-routed motion adapters. An optional control branch further incorporates available motion descriptors for controllable synthesis. Experiments on ACDC, M\&Ms, and M\&Ms-2 demonstrate consistent improvements in geometric accuracy and functional fidelity. Under ED-only synthesis, our method achieves biventricular ASSD, HD95, and vRMSE of \(1.49\pm0.34\)~mm, \(3.77\pm1.06\)~mm, and \(3.31\pm1.03\)~mm, respectively, outperforming all competing methods. Complementary functional and robustness evaluations further demonstrate that the synthesized sequences preserve physiologically plausible ventricular dynamics and generalize across cohorts and disease phenotypes. The code will be released publicly upon acceptance of the manuscript for publication.

\end{abstract}


\end{frontmatter}

\section{Introduction}

Cardiovascular disease is a major cause of death worldwide \citep{Journal/JACC/Roth2019}.
Cardiac abnormalities are reflected not only in changes in anatomy, but also in altered functional and motion patterns throughout the cardiac cycle \citep{Journal/MIA/Bai2015,Journal/MIA/Puyol2017}. 
Cardiac magnetic resonance imaging (MRI) can characterize myocardial shape and motion during contraction and relaxation with high spatial and temporal resolution \citep{Journal/Plos/Wan2015}. 
However, analyzing such spatiotemporal information only in the voxel or label space makes it difficult to directly represent continuous anatomical surfaces, topological structures, and geometrical changes that are aligned across time. 
In contrast, time-resolved 3D mesh sequences provide a more natural representation of cardiac surface geometry and temporal correspondence, and offer a useful basis for patient-specific visualization and physics-based computational modeling \citep{Journal/TMI/Meng2023,Journal/TMI/Kong2022,Journal/COIP/Marsden2015}.

Despite these advantages, complete and high-quality 4D (3D+t) cardiac mesh sequences remain difficult to obtain in practice \citep{Conference/ICCV/Yuan2023,Conference/EMBC/Banerjee2022}, and in real clinical settings only a few key states can usually be acquired reliably. 
Previous studies \citep{Journal/Arix/Liu2026,Journal/MIA/Kong2021,Journal/MIA/Luo2026,Journal/Arxiv/Chen2026} have shown that even under very limited observations, such as a single frame, sparse slices, or a reference phase, it is still possible to recover patient-specific cardiac dynamics. 
Among these states, end-diastole (ED) is commonly used as a reference because it provides a relatively stable and complete anatomical configuration and serves as an anchor for subsequent dynamic modeling \citep{Journal/MIA/Zakeri2023,Journal/TMI/Meng2023,Conference/EMBC/Upendra2021}. 
These observations motivate using the ED anatomy as an accessible patient-specific condition for synthesizing full-cycle cardiac dynamics.

Generative modeling has become increasingly valuable in cardiac analysis, not only for completing missing dynamics, but also for expanding the available sample space, constructing virtual cohorts, and supporting in silico clinical trials \citep{Journal/PTRS/Niederer2020,Conference/MICCAI/Dou2023,Conference/MICCAI/Sorensen2024,Conference/MICCAI/Dou2025}.
Early efforts have largely focused on static cardiac anatomy modeling \citep{Journal/MIA/Kong2021,Journal/FCM/Beetz2022,Conference/MICCAI/Dou2023,Journal/CMIG/Beetz2025}, including patient-specific mesh reconstruction from cardiac images and conditional generation of anatomical shapes or virtual populations. 
These methods provide important tools for characterizing inter-subject morphological variability. 
However, static anatomy alone cannot describe functional changes over the cardiac cycle, nor can it capture the coupled evolution of shape and motion that underlies cardiac function.
Recent studies have increasingly shifted toward dynamic cardiac modeling \citep{Journal/TMI/Qiao2023,Conference/MICCAI/Sorensen2024,Conference/MICCAI/Dou2025,Journal/NMI/Qiao2025}, enabling tasks such as motion sequence completion, dynamic shape generation, and joint shape-motion distribution modeling. 
Despite these advances, many existing dynamic generative models mainly emphasize population-level plausibility and temporal consistency, while explicit modeling of regional motion variation remains limited. 
This is particularly important because cardiac motion is spatially heterogeneous, with regional wall motion differences reflecting local myocardial function and pathology \citep{Journal/MIA/Bai2015,Conference/MICCAI/Xue2018}. 
Without explicitly accounting for such regional characteristics, dynamic generation models may struggle to represent localized abnormalities and fine-grained functional variation.

A second challenge is modeling phenotype-dependent motion variability.
Many existing dynamic generation methods are trained primarily on healthy subjects or large normative cohorts \citep{Journal/NMI/Qiao2025,Conference/MICCAI/Dou2025}, which may limit their ability to represent pathological motion patterns. 
Although several studies have introduced conditional control for cardiac sequence generation \citep{Journal/TMI/Qiao2023,Conference/MICCAI/Sorensen2024,Conference/MICCAI/Dou2025}, the conditioning signals are often restricted to general clinical covariates or diagnostic labels, and thus may not fully capture fine-grained, patient-specific functional differences. 
Clinically meaningful cardiac motion generation therefore requires more informative conditioning mechanisms, particularly phenotype-aware representations that can encode individual disease characteristics and guide the generation of region-specific dynamic patterns.

To address the above challenges, we propose a region-specific and phenotype-adaptive latent generative framework for full-cycle biventricular motion synthesis on unified-topology meshes. The framework uses a patient-specific ED mesh as the anatomical condition and synthesizes full-cycle cardiac dynamics under phenotype guidance, aiming to capture anatomy-dependent and disease-associated motion variation. Specifically, our method first learns a motion-informed functional partition from real cardiac sequences, providing explicit regional priors for modeling local dynamics. The priors are then used to organize a motion variational autoencoder (VAE) for reconstructing full-cycle cardiac sequences on the learned regional structure. Finally, a rectified-flow generator is trained in the learned VAE latent space to enhance phenotype-conditioned motion synthesis from ED anatomy, with optional motion descriptors used when available for controllable generation.
The main contributions of this work are as follows:

\begin{enumerate}[label=\roman*.]
    \item We establish a phenotype-adaptive latent motion generation framework for synthesizing full-cycle biventricular motion from a single ED mesh.
    \item We introduce a motion-informed regional prior and integrate it into a region-structured motion VAE, enabling compact representation of full-cycle cardiac dynamics while preserving regional motion heterogeneity.
    \item We develop a rectified-flow latent generator with prototype-routed motion adapters, improving phenotype-conditioned motion generation while supporting optional functional control.
    \item Extensive experiments on multiple public cardiac MRI datasets demonstrate the effectiveness of the proposed method in full-cycle motion synthesis, functional consistency, and cross-disease generalization. 
\end{enumerate}

This paper significantly extends a preliminary conference version of the work presented in \citet{Conference/MICCAI/Yang2026}. Compared with the previous framework, this study redesigns the overall pipeline into a region-specific latent generative framework for full-cycle motion synthesis, strengthening both the structural prior and the latent generation mechanism. Specifically, we replace the fixed binary regional adjacency with a hop-based region-topology prior, improving the modeling of gradual motion propagation across neighboring regions. We also replace the previous VAE-prior-based motion generation with a rectified-flow latent generator in the learned latent space, where fine-grained ED anatomical and phenotype conditioning support more detailed morphology-aware motion synthesis. In addition, an optional functional-control branch enables controllable modulation of the generated dynamics when motion descriptors are available. We further conduct more systematic experiments, including comparisons with competing methods and ablation studies, to evaluate the contribution of these modeling components.

\section{Related Work} \label{related_work}

\subsection{Cardiac Shape and Motion Generative Modeling}
Cardiac statistical shape analysis has  been used to characterize anatomical variation and its relationship to disease, supporting diagnosis, risk assessment, and treatment planning \citep{Journal/MIA/Bai2015,Journal/Heart/Biglino2017,Journal/NM/Bai2020}. With the development of deep learning, research has gradually shifted toward more expressive generative models. For instance, \citet{Conference/MICCAI/Dou2023} proposed a conditional VAE to synthesize virtual anatomical populations of the left ventricle under covariate constraints. \citet{Conference/STACOM/Beetz2021,Journal/CMIG/Beetz2025} used point clouds and VAE to model three-dimensional cardiac shape variation, and applied it to myocardial infarction-related shape analysis and virtual heart synthesis. \citet{Journal/MIA/Kong2024} further extended conditional anatomical generation to more complex topologies, enabling controllable synthesis of congenital heart disease anatomy through type-shape disentanglement. Overall, static cardiac mesh generation has evolved from patient-specific anatomical reconstruction to controllable anatomical distribution learning. However, these methods remain limited to static anatomy and cannot directly describe functional changes throughout the cardiac cycle. 
To solve this, \citet{Journal/TMI/Qiao2023} proposed a conditional spatiotemporal generative model to jointly model 4D cardiac anatomy and its relationship with non-imaging clinical factors. \citet{Journal/NMI/Qiao2025} directly learned the shape-motion distribution of 3D+t biventricular meshes and explored cardiac morphology and motion patterns in large-scale population data. \citet{Conference/MICCAI/Dou2025} further generated dynamic virtual heart populations through spatiotemporal disentanglement. Nevertheless, existing dynamic models mainly focus on globally coherent motion patterns and are often developed on healthy populations or under global condition signals, leaving disease-related dynamic variation largely underexplored.

\subsection{Patient-Specific Dynamic Modeling from Sparse Observations}
Patient-specific cardiac dynamic modeling has mainly relied on registration, motion tracking, and spatiotemporal deformation estimation from cine imaging \citep{Journal/MP/Qiao2020,Conference/CVPR/Ye2021,Journal/CIBM/Lu2023,Journal/MIA/Qin2023,Journal/MIA/Lopez2023,Conference/NIPS/Yang2024}. With the development of deep learning, these approaches have gradually shifted from traditional optimization-based registration to data-driven dynamic recovery. One group of methods \citep{Journal/MIA/Qin2023,Journal/MIA/Zakeri2023,Journal/CIBM/Lu2023}, learns temporal deformation representations in the image domain. For example, \citet{Journal/MIA/Qin2023} achieved left ventricular myocardial motion tracking through a biomechanically constrained latent deformation manifold. DragNet \citep{Journal/MIA/Zakeri2023} further recovered temporally consistent full cardiac dynamics from a single-frame cine MRI. However, the outputs of these methods are still essentially image-domain spatiotemporal deformations or generated image sequences, rather than mesh sequences with unified topological constraints.

To obtain more explicit geometric representations, recent studies moved from image-domain displacement fields to mesh-based modeling. DeepMesh \citep{Journal/TMI/Meng2023} represented patient-specific cardiac anatomy by deforming a template mesh to the ED geometry and subsequently estimated three-dimensional mesh motion. \citet{Journal/MIA/Laumer2023} further recovered patient-specific 4D cardiac meshes from 2D echocardiographic videos under a weakly supervised setting. Compared with general image-domain motion fields, mesh sequences can more naturally preserve surface geometry, fixed topology, and vertex correspondence across time, making them more suitable for subsequent regional prior modeling and local dynamic analysis. Overall, existing patient-specific dynamic modeling studies have shown that recovering full cardiac dynamics from key states, single-frame images, or limited observations is a reasonable and important problem setting. However, their main goal is still tracking, reconstruction, or globally plausible motion recovery, rather than unified generative modeling of disease-related dynamic processes.

\subsection{Latent and Flow-Based Motion Generation}
In complex temporal motion modeling, latent-variable generative models are among the most representative technical routes \citep{Conference/Arxiv/Kingma2013,Conference/NIPS/Chung2015,Conference/NIPS/Sohn2015}. Their core idea is to compress high-dimensional temporal motion into a low-dimensional latent space and then model complex motion distributions within that space. Action2Motion  \citep{Conference/ACMMM/Guo2020} used a temporal VAE to model human motion under action conditions. ACTOR \citep{Conference/ICCV/Petrovich2021} further employed a Transformer to learn action-aware latent motion representations. Furthermore, AnimateAnyMesh \citep{Conference/ICCV/Wu2025} extended conditional generation in compressed latent spaces to three-dimensional mesh objects with arbitrary topology, enabling unified animation generation for multiple types of mesh objects. This suggests that latent-space modeling is not only suitable for general temporal signals, but can also support motion generation for geometrically more complex and topologically more flexible 3D objects. 

Diffusion models \citep{Conference/NIPS/Ho2020,Conference/NIPS/Dhariwal2021,Conference/Arxiv/Song2020} achieve high-fidelity generation through iterative denoising and have shown strong performance in modeling complex distributions, but they usually rely on multi-step sampling and large training scales. To improve training and sampling efficiency, recent studies have turned to continuous flow-based modeling. Flow Matching \citep{Conference/Arxiv/Lipman2022} characterizes transport between distributions by learning continuous-time velocity fields along fixed probability paths. Rectified Flow \citep{Conference/Arxiv/Liu2022} further favors learning more direct transport trajectories. Motion Flow Matching \citep{Conference/Arxiv/Hu2023} has applied this idea to human motion generation, completion, and interpolation. However, these methods have mostly been developed for general human motion or generic 3D geometric object scenarios, and they still lack specialized designs for jointly encoding anatomical states, disease-related conditions, and regional priors. Therefore, combining latent-space modeling with flow-based generation, and adapting it to disease-related and regionally heterogeneous cardiac dynamics, remains an important direction for further study.


\section{Methods} \label{method}


\subsection{Problem Formulation}
\label{sec:method:problem}

\begin{figure}[t]
    \centering
    \includegraphics[
        width=\columnwidth,
        keepaspectratio
    ]{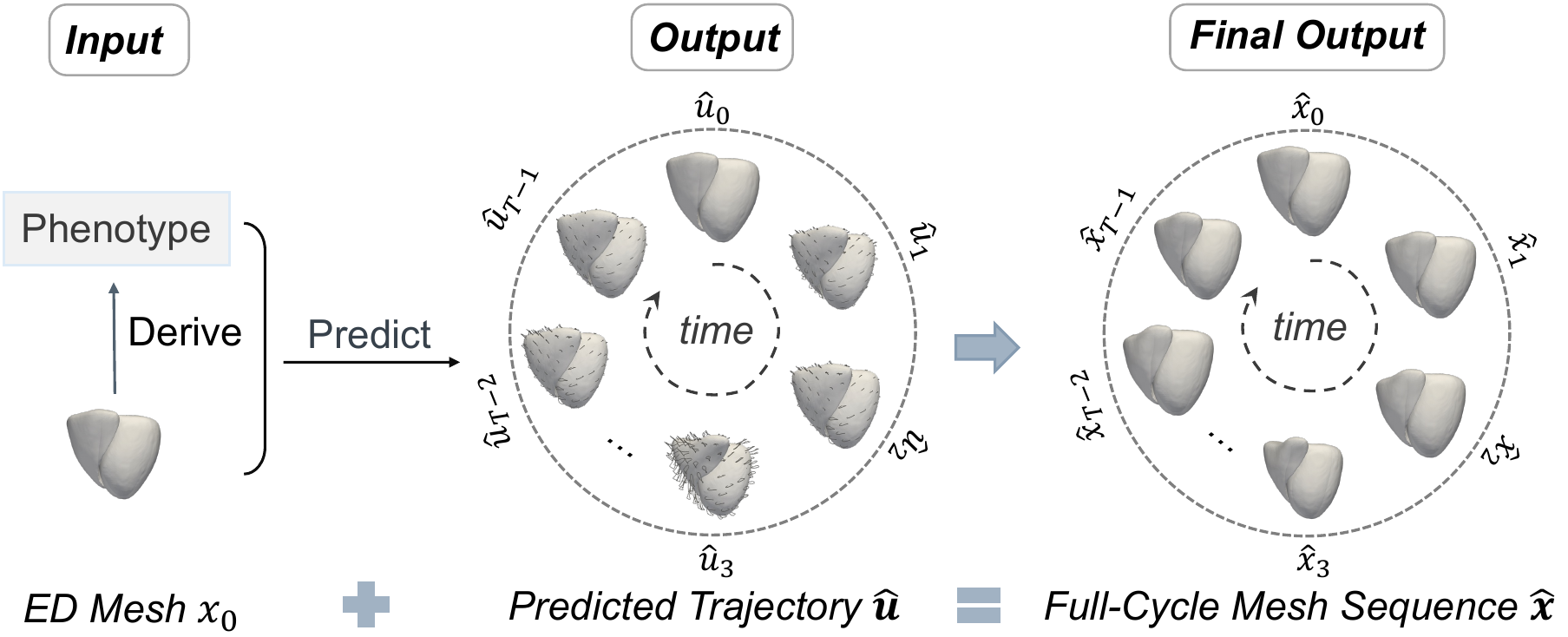}
    \caption{Illustration of the task and its formulation. ED: end-diastolic.}
    \label{fig:task}
\end{figure}

Given only the ED biventricular mesh of a subject, our goal is to recover the full dynamic biventricular mesh sequence. 
Let
$x_t \in \mathbb{R}^{N\times 3}$ denote the biventricular mesh at cardiac frame $t$, where $N$ is the number of vertices and $t=0,\ldots,T-1$. 
The full cardiac mesh sequence is defined as $\mathbf{x}=\{x_t\}_{t=0}^{T-1}\in\mathbb{R}^{T\times N\times 3}$, where $x_0$ corresponds to the ED mesh.
All biventricular meshes are represented using a unified template with shared topology and vertex-wise correspondence across subjects and cardiac frames. 
This allows cardiac motion to be formulated as an ED-referenced displacement trajectory, reducing subject-specific spatial variation and enabling the model to focus on learning cardiac deformation rather than absolute vertex coordinates. 
Specifically, the displacement at frame $t$ is defined as $u_t = x_t - x_0$.
The complete displacement trajectory is therefore given by $\mathbf{u}=\{u_t\}_{t=0}^{T-1}\in\mathbb{R}^{T\times N\times 3}$.
Under this formulation, cardiac sequence recovery is defined as learning a mapping from the patient-specific ED mesh $x_0$ to the full displacement trajectory $\hat{\mathbf{u}}$, as shown in Fig.~\ref{fig:task}. 
The dynamic mesh sequence is then recovered by adding the generated displacement field to the ED mesh:
\begin{equation}
    \hat{x}_t = x_0+\hat{u}_t,
    \qquad
    t=0,\ldots,T-1.
\end{equation}
To predict the displacement trajectory \(\hat{\mathbf{u}}\), we learn a motion generator conditioned on the patient-specific ED mesh \(x_0\) and its ED-derived phenotype descriptors \(c_{ED}^{global}\) and \(c_{ED}^{fine}\).
Following the cardiac MRI-derived phenotype definitions in~\citet{Journal/NM/Bai2020}, \(c_{ED}^{global}\) summarizes global chamber and myocardial morphology, whereas \(c_{ED}^{fine}\) further captures regional ED myocardial wall-thickness information (see Sec.~\ref{sec:data} for details).
When ventricular functional measurements are available, the motion descriptor \(c^{motion}\) can be additionally incorporated as optional inputs.

\begin{figure*}[t]\center
 \includegraphics[width=0.90\textwidth]{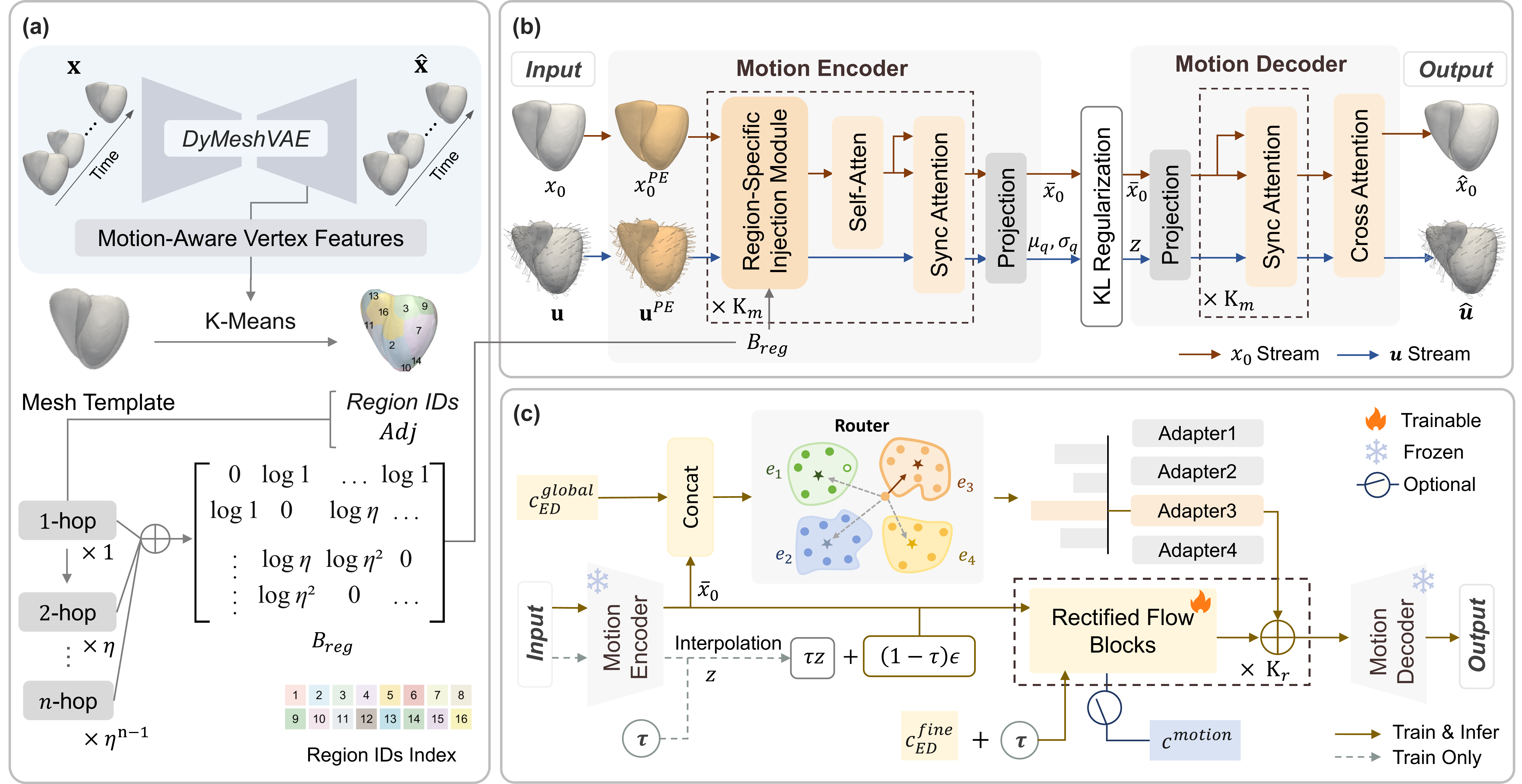}\\[-2ex]
   \caption{Illustration of the proposed regional motion-informed and phenotype-guided biventricular motion synthesis framework. 
   (a) Motion-driven functional regionalization.
   (b) Region-topology aware motion variational autoencoder (VAE).
   (c) Phenotype-guided latent flow for anatomy-to-motion generation.
   DyMeshVAE: dynamic mesh VAE; PE: positional encoding.}
\label{fig:method:framework}
\end{figure*}

\subsection{Motion-Informed Cardiac Functional Parcellation} \label{sec:method:functional parcellation}

To model spatially heterogeneous cardiac deformation, we learn a data-driven functional parcellation directly from training motion sequences, avoiding predefined anatomical divisions that may not reflect regional motion similarity.
Defined on the unified biventricular template, the resulting partition provides a fixed, subject-independent regional prior.
Specifically, given full cardiac mesh sequences \(\mathbf{x}\), we first train a dynamic mesh reconstruction network, DyMeshVAE~\citep{Conference/ICCV/Wu2025}, to extract vertex-wise motion features \(f(i)\), as shown in Fig.~\ref{fig:method:framework} (a). 
To obtain a subject-independent descriptor for each template vertex, we average the vertex features across the training set and thus obtain \(\bar{f}(i)\). 
We then apply K-means to the template-level descriptors and partition the template vertices into \(N_R\) motion regions:
\begin{equation}
r(i)
=
\arg\min_{\rho\in\{1,\ldots,N_R\}}
\left\|
\bar{f}(i)-\mu_\rho
\right\|_2^2,
\qquad
r(i)\in\{1,\ldots,N_R\},
\label{eq:region_assignment}
\end{equation}
where \(i\) is the index of mesh vertex and \(\mu_\rho\) is the centroid of the \(\rho\)-th motion region.
The resulting regional prior comprises the fixed vertex-to-region assignments \(\{r(i)\}_{i=1}^{N}\) and the region adjacency matrix \(Adj\). Two regions are considered adjacent if at least one mesh edge connects vertices assigned to the two regions.

Cardiac motion involves coordinated deformation across multiple interconnected regions, which may not be fully captured by one-hop connectivity alone. We therefore introduce multi-hop regional attention bias (MRAB) to extend the binary adjacency mask used in ~\citet{Conference/MICCAI/Yang2026} by encoding the multi-hop structure of the learned region graph in the attention logits.
Let \(D_{\mathrm{reg}}(\rho,\rho')\) denote the shortest-path distance between regions \(\rho\) and \(\rho'\).
Rather than treating all admissible region pairs equally, MRAB converts this distance into an additive attention bias:
\begin{equation}
B_{\mathrm{reg}}(\rho,\rho')
=
\begin{cases}
0,
& D_{\mathrm{reg}}(\rho,\rho')=0,\\
\bigl(D_{\mathrm{reg}}(\rho,\rho')-1\bigr)\log\eta,
& 1\leq D_{\mathrm{reg}}(\rho,\rho')
\leq L_{\mathrm{hop}},\\
-\infty,
& D_{\mathrm{reg}}(\rho,\rho')>L_{\mathrm{hop}},
\end{cases}
\label{eq:hop_bias}
\end{equation}
where \(0<\eta<1\) controls the attenuation across successive hops and \(L_{\mathrm{hop}}\) specifies the maximum interaction range.
Accordingly, self- and one-hop interactions remain unbiased, whereas every additional hop introduces a progressively stronger negative bias. Region pairs outside the prescribed neighborhood are excluded.
MRAB thus provides a distance-aware topological prior for controlled information exchange beyond immediately adjacent regions. The vertex-to-region assignments and \(B_{\mathrm{reg}}\) are subsequently used for regional feature aggregation and topology-aware attention, respectively, in the motion VAE described in Sec.~\ref{sec:method:motionVAE}.

\subsection{Region-Structured Motion VAE} \label{sec:method:motionVAE}

\begin{figure}[t]
    \centering
    \includegraphics[width=0.88\linewidth]{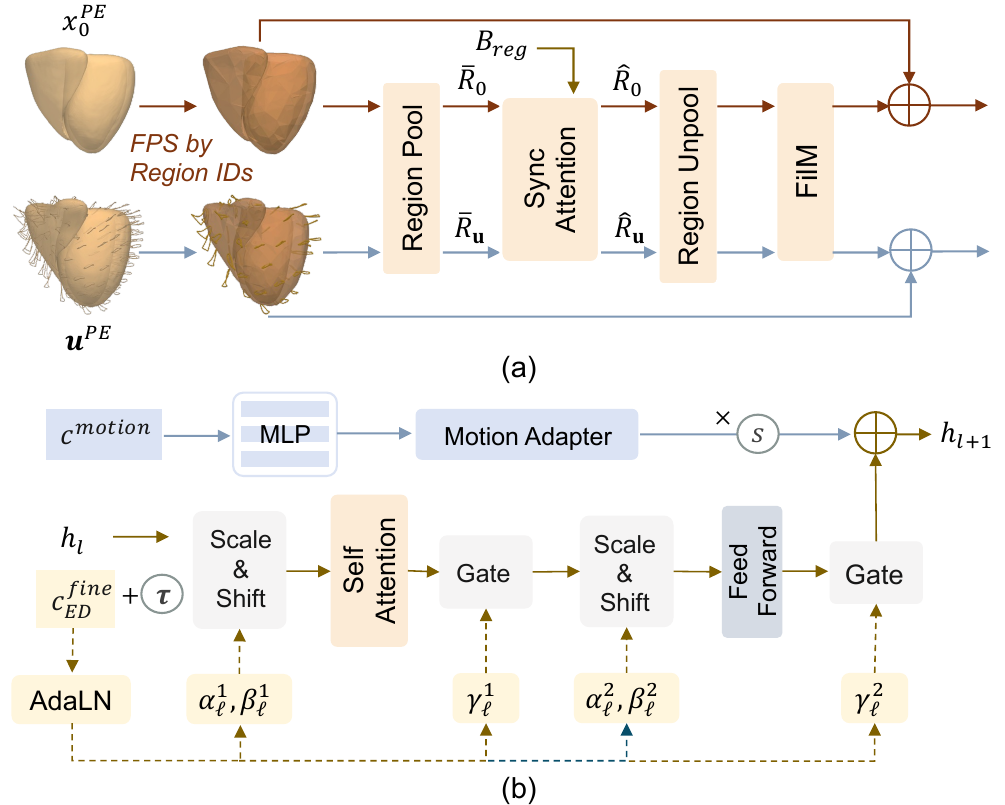}
    \caption{
    (a) Region-specific injection module.
    (b) Phenotype-conditioned rectified-flow block with optional functional motion control. FPS: farthest point sampling; MLP: multilayer perceptron. }
    \label{fig:region_injection}
\end{figure}
To learn a compact latent representation for full-cycle cardiac motion, we design a region-structured motion VAE, as illustrated in Fig.~\ref{fig:method:framework} (b).
The VAE adopts a two-stream design in which the ED stream encodes the static anatomical state \(x_0\), while the motion stream encodes the ED-relative trajectory \(\mathbf{u}\). This separation allows the motion stream to model temporal deformation relative to a stable anatomical reference, reducing interference between static anatomical coordinates and temporal displacement patterns.
Inspired by DyMeshVAE~\citep{Conference/ICCV/Wu2025}, we apply distinct positional encoding (PE) to \(x_0\) and \(\mathbf{u}\), yielding \(x_0^{PE}\) and \(\mathbf{u}^{{PE}}\), respectively. Region-balanced farthest point sampling (FPS) then selects \(N_a\) anchor vertices, and the same sampling indices are used to gather features from both streams, forming paired ED and motion anchor tokens \(a_0,a_u\in\mathbb{R}^{N_a\times d}\), where \(d\) denotes the token feature dimension. Each token pair inherits the region label of its sampled vertex from the motion-informed partition defined in Sec.~\ref{sec:method:functional parcellation}.

To preserve the regional motion heterogeneity captured by this partition, each encoder block incorporates a region-specific injection module, as shown in Fig.~\ref{fig:region_injection} (a). The module mean-pools the anchor tokens within each region to obtain ED and motion region tokens \(\bar R_0,\bar R_{\mathbf u}\in\mathbb{R}^{N_R\times d}\).
At the regional level, the two streams retain distinct roles. The ED region tokens determine how information is exchanged between regions, whereas the motion region tokens provide the dynamic information propagated through these interactions. Accordingly, \(B_{\mathrm{reg}}\) is added to the attention logits computed from \(\bar R_0\), and the resulting attention map is applied to both streams:
\begin{equation}
\begin{aligned}
\hat{R}_{0}
&=
\mathrm{Softmax}
\left(
\frac{\bar{R}_{0}\bar{R}_{0}^{\top}}{\sqrt{d}}
+
B_{\mathrm{reg}}
\right)
\bar{R}_{0}
+
\bar{R}_{0},\\
\hat{R}_{\mathbf{u}}
&=
\mathrm{Softmax}
\left(
\frac{\bar{R}_{0}\bar{R}_{0}^{\top}}{\sqrt{d}}
+
B_{\mathrm{reg}}
\right)
\bar{R}_{\mathbf{u}}
+
\bar{R}_{\mathbf{u}}.
\end{aligned}
\label{eq:region_sync}
\end{equation}
The updated region tokens are then mapped back to their corresponding anchors and injected into both streams through FiLM-based residual modulation~\citep{Conference/AAAI/Perez2018}. Stacking \(K_m\) such encoder blocks progressively incorporates topology-constrained regional context into the anchor features while preserving fine-grained spatial representations.

After the final encoder block, the ED anchor tokens are projected into the deterministic anatomical latent \(\bar{x}_0\in\mathbb{R}^{N_a\times d_z}\), while the motion anchor tokens are projected into \(\mu_q\) and \(\sigma_q\), which parameterize the motion posterior.

\begin{equation}
q_\phi(z\mid x_0,\mathbf{u})
=
\mathcal{N}
\left(
\mu_q,
\mathrm{diag}(\sigma_q^2)
\right),
\label{eq:vae_posterior}
\end{equation}
where \(z\in\mathbb{R}^{N_a\times d_z}\) denotes a motion latent sampled from the posterior and \(d_z\) is the per-anchor latent feature dimension.
The decoder takes the deterministic anatomical latent \(\bar{x}_0\) and motion latent \(z\) as inputs, projects them back to the ED and motion anchor representations, and refines both streams through SyncAttention. The dense ED features \(x_0^{PE}\) then query the decoded ED anchor representations through cross-attention and aggregate the corresponding motion anchor representations to reconstruct \(\hat{\mathbf{u}}\in\mathbb{R}^{T\times N\times3}\).
We define the reconstruction term as the mean squared error between the reconstructed and reference dynamic mesh sequences:
\begin{equation}
\mathcal{L}_{rec}
=
\frac{1}{T}\frac{1}{N}
\sum_{t=0}^{T-1}
\sum_{i=1}^{N}
\left\|
\hat{x}_t(i)-x_t(i)
\right\|_2^2 .
\label{eq:vae_rec_loss}
\end{equation}
The motion VAE is optimized with dynamic mesh reconstruction and KL regularization:
\begin{equation}
\mathcal{L}_{VAE}
=
\mathcal{L}_{rec}
+
\lambda_{KL}
\mathcal{D}_{KL}
\left[
q_\phi(z\mid x_0,\mathbf{u})
\|
\mathcal{N}(0,I)
\right],
\label{eq:vae_loss}
\end{equation}
where \(\lambda_{KL}\) is a balancing parameter.
After training, the motion VAE is frozen, providing the latent space and decoder used by the phenotype-guided motion generator described in Sec.~\ref{sec:method:motion_gen}.

\subsection{Phenotype-Guided Latent Motion Generation}
\label{sec:method:motion_gen}

\paragraph{Latent Rectified-Flow Training}
As described in Sec.~\ref{sec:method:motionVAE}, the frozen motion VAE represents each training sequence using a deterministic anatomical latent \(\bar{x}_0\) and a posterior motion latent \(z\), from which its decoder reconstructs the displacement trajectory \(\mathbf{u}\). Its Gaussian prior, however, is independent of subject-specific ED anatomy and phenotype. We therefore train a conditional rectified-flow (RF) model whose velocity field is conditioned on \(\bar{x}_0\), \(c_{ED}^{global}\), and \(c_{ED}^{fine}\), thereby transporting Gaussian noise into subject-specific motion latents, as illustrated in Fig.~\ref{fig:method:framework} (c).
Following the standard RF formulation~\citep{Conference/Arxiv/Liu2022,Conference/Arxiv/Lipman2022}, for each motion latent \(z\) sampled from the frozen VAE posterior, we draw \(\epsilon\sim\mathcal{N}(0,I)\) and \(\tau\sim\mathcal{U}(0,1)\), and construct \(z_\tau=(1-\tau)\epsilon+\tau z\) with target velocity \(z-\epsilon\). The velocity network comprises \(K_r\) RF blocks, as shown in Fig.~\ref{fig:region_injection} (b). The representations of \(z_\tau\) and \(\bar{x}_0\) are summed to initialize the hidden representation. Within each block, \(c_{ED}^{fine}\) and \(\tau\) are incorporated through AdaLN~\citep{Conference/ICCV/Peebles2023,Conference/ICLR/Yang2025}, whereas \(\bar{x}_0\) and \(c_{ED}^{global}\) jointly determine the prototype-specific adapter route described below.
The RF model is optimized by
\begin{equation}
\mathcal{L}_{RF}
=
\mathbb{E}_{z,\epsilon,\tau}
\left[
\left\|
v_\theta
\left(
z_\tau,
\bar{x}_0,
c_{ED}^{global},
c_{ED}^{fine},
\tau
\right)
-
(z-\epsilon)
\right\|_2^2
\right].
\label{eq:rf_loss}
\end{equation}

\paragraph{Prototype-Routed Motion Adapters}
A shared RF backbone models motion patterns common across subjects.
To adapt this shared velocity field to heterogeneous ED anatomy and phenotypes without training separate generators, we introduce \(N_e\) lightweight motion adapters, each associated with one routing prototype.
For each subject, we form the routing feature \(g_{\mathrm{route}}\) by concatenating the mean-pooled anatomical latent \(\mathrm{MeanPool}(\bar{x}_0)\) with the global phenotype descriptor \(c_{ED}^{global}\). Applying K-means to the training routing features yields \(N_e\) routing prototypes \(\{e_j\}_{j=1}^{N_e}\). Each subject is assigned to its nearest routing prototype:
\begin{equation}
j^*
=
\arg\min_{j\in\{1,\ldots,N_e\}}
\left\|
g_{\mathrm{route}}-e_j
\right\|_2^2,
\label{eq:route_assignment}
\end{equation}
where \(e_j\) denotes the \(j\)-th routing prototype. The selected index \(j^*\) activates the corresponding adapter in each RF block, allowing the shared velocity field to specialize according to subject-specific ED anatomy and global phenotype.

\paragraph{Optional Motion Control}
ED anatomy does not completely determine functional properties such as contraction strength and motion amplitude.
When the motion descriptor \(c^{motion}\) is available, it is encoded by a multilayer perceptron (MLP) and injected into each RF block through an additional Motion Adapter.
As shown in Fig.~\ref{fig:region_injection} (b), the resulting residual is scaled by the control scale \(s\) and added to the hidden representation:
\begin{equation}
h_{\ell+1}
\leftarrow
h_{\ell+1}
+
s\,
\mathcal{M}_{\ell}
\left(
\mathrm{MLP}(c^{motion})
\right),
\label{eq:motion_control_adapter}
\end{equation}
where \(h_{\ell+1}\) denotes the output hidden representation of the \(\ell\)-th RF block and \(\mathcal{M}_\ell\) denotes its Motion Adapter.
The coefficient \(s\) controls the strength of functional modulation.

\paragraph{Inference}
Given an ED mesh \(x_0\), the frozen ED encoder produces the anatomical latent \(\bar{x}_0\). The routing feature \(g_{\mathrm{route}}\) is then constructed from \(\bar{x}_0\) and \(c_{ED}^{global}\), and Eq.~\eqref{eq:route_assignment} selects the adapter route \(j^*\), which remains fixed throughout RF integration. Starting from \(z_{\tau=0}=\epsilon\), where \(\epsilon\sim\mathcal{N}(0,I)\), the terminal motion latent is obtained by
\begin{equation}
\hat{z}
=
z_{\tau=1}
=
\epsilon
+
\int_0^1
v_\theta^{(j^*)}
\left(
z_\tau,
\bar{x}_0,
c_{ED}^{fine},
\tau
\right)
\,\mathrm{d}\tau,
\label{eq:inference}
\end{equation}
where \(v_\theta^{(j^*)}\) denotes the RF velocity field with the \(j^*\)-th adapter route activated. When \(c^{motion}\) is available, the motion adapter additionally modulates the velocity field throughout integration; otherwise, this branch is disabled. Finally, the terminal latent \(\hat{z}=z_{\tau=1}\), together with \(\bar{x}_0\), is passed to the frozen motion decoder to reconstruct the full-cycle ED-relative displacement trajectory \(\hat{\mathbf{u}}\).

\begin{table}[t]
\centering
\caption{Composition of the study cohort across datasets and diagnostic groups. NOR: normal; DCM: dilated cardiomyopathy; HCM: hypertrophic cardiomyopathy; RVA: right-ventricular abnormality.}
\label{exp:tb:dataset}
\footnotesize
\begin{tabular}{lccccc}
\toprule
Dataset & NOR & DCM & HCM & RVA & Total \\
\midrule
ACDC & 30 & 30 & 30 & 30 & 120 \\
M\&Ms & 89 & 97 & 85 & 15 & 286 \\
M\&Ms-2 & 75 & 60 & 60 & 65 & 260 \\
\midrule
Total & 194 & 187 & 175 & 110 & 666 \\
\bottomrule
\end{tabular}
\end{table}

\section{Experiments and Results} 

\begin{table*}[htbp]
\centering
\caption{Quantitative comparison under the ED-conditioned full-cycle biventricular motion synthesis.
ASSD: average symmetric surface distance; HD95: 95th percentile Hausdorff distance; vRMSE: vertex-wise root mean square error;
BiV: biventricular; LV: left ventricular; RV: right ventricular. Values are reported as the mean \(\pm\) standard deviation across test subjects. For each subject, each metric was first averaged across 20 independently generated motion sequences.}
\label{tab:main_comparison}

\footnotesize
\setlength{\tabcolsep}{2.5pt}
\renewcommand{\arraystretch}{1.1}

\resizebox{\textwidth}{!}{%
\begin{tabular}{@{}lccccccccc@{}}
\toprule
\multirow{2}{*}{Method}
& \multicolumn{3}{c}{ASSD (mm) $\downarrow$} 
& \multicolumn{3}{c}{HD95 (mm) $\downarrow$} 
& \multicolumn{3}{c}{vRMSE (mm) $\downarrow$} \\
\cmidrule(lr){2-4}
\cmidrule(lr){5-7}
\cmidrule(lr){8-10}
& LV & RV & BiV 
& LV & RV & BiV 
& LV & RV & BiV \\
\midrule

CVAE~\citep{Conference/NIPS/Sohn2015}
& \(2.16\pm0.66\) & \(2.08\pm0.63\) & \(2.15\pm0.54\)
& \(4.34\pm1.30\) & \(4.78\pm1.77\) & \(4.71\pm1.54\)
& \(3.86\pm1.34\) & \(4.25\pm1.54\) & \(4.11\pm1.25\) \\

ACTOR~\citep{Conference/ICCV/Petrovich2021}
& \(2.10\pm0.64\) & \(2.00\pm0.80\) & \(2.11\pm0.60\)
& \(4.28\pm1.41\) & \(4.85\pm2.33\) & \(4.93\pm1.98\)
& \(3.71\pm1.39\) & \(4.14\pm1.83\) & \(4.01\pm1.52\) \\

Action2Motion~\citep{Conference/ACMMM/Guo2020}
& \(2.28\pm0.71\) & \(2.18\pm0.78\) & \(2.22\pm0.64\)
& \(4.81\pm1.65\) & \(5.07\pm2.18\) & \(5.16\pm1.98\)
& \(4.20\pm1.53\) & \(4.51\pm1.85\) & \(4.21\pm1.52\) \\

CHeart~\citep{Journal/TMI/Qiao2023}
& \(2.21\pm0.65\) & \(2.04\pm0.72\) & \(2.09\pm0.53\)
& \(4.48\pm1.43\) & \(4.73\pm1.98\) & \(4.97\pm1.52\)
& \(3.80\pm1.37\) & \(4.10\pm1.70\) & \(3.98\pm1.22\) \\

MeshHeart~\citep{Journal/NMI/Qiao2025}
& \(2.09\pm0.65\) & \(2.01\pm0.63\) & \(2.08\pm0.53\)
& \(4.23\pm1.30\) & \(4.64\pm1.77\) & \(4.68\pm1.54\)
& \(3.78\pm1.34\) & \(4.16\pm1.54\) & \(4.13\pm1.25\) \\

4DCardioSynth~\citep{Conference/MICCAI/Dou2025}
& \(2.07\pm0.61\) & \(2.04\pm0.65\) & \(2.12\pm0.55\)
& \(4.22\pm1.29\) & \(4.73\pm1.72\) & \(4.76\pm1.70\)
& \(3.63\pm1.25\) & \(4.11\pm1.50\) & \(3.94\pm1.33\) \\

RePCM~\citep{Conference/MICCAI/Yang2026}
& \(1.79\pm0.60\) & \(2.00\pm0.76\) & \(1.69\pm0.36\)
& \(3.94\pm1.48\) & \(4.54\pm2.04\) & \(4.27\pm1.83\)
& \(3.43\pm1.42\) & \(4.08\pm1.74\) & \(3.61\pm1.45\) \\

\midrule

\textbf{Ours}
& \(\mathbf{1.63\pm0.48}\) 
& \(\mathbf{1.85\pm0.50}\) 
& \(\mathbf{1.49\pm0.34}\)
& \(\mathbf{3.56\pm1.09}\) 
& \(\mathbf{4.24\pm1.23}\) 
& \(\mathbf{3.77\pm1.06}\)
& \(\mathbf{3.13\pm1.09}\) 
& \(\mathbf{3.74\pm1.12}\) 
& \(\mathbf{3.31\pm1.03}\) \\

\bottomrule
\end{tabular}%
}

\end{table*}

\begin{figure*}[t]
    \centering
    \includegraphics[
        width=0.95\textwidth,
        keepaspectratio
    ]{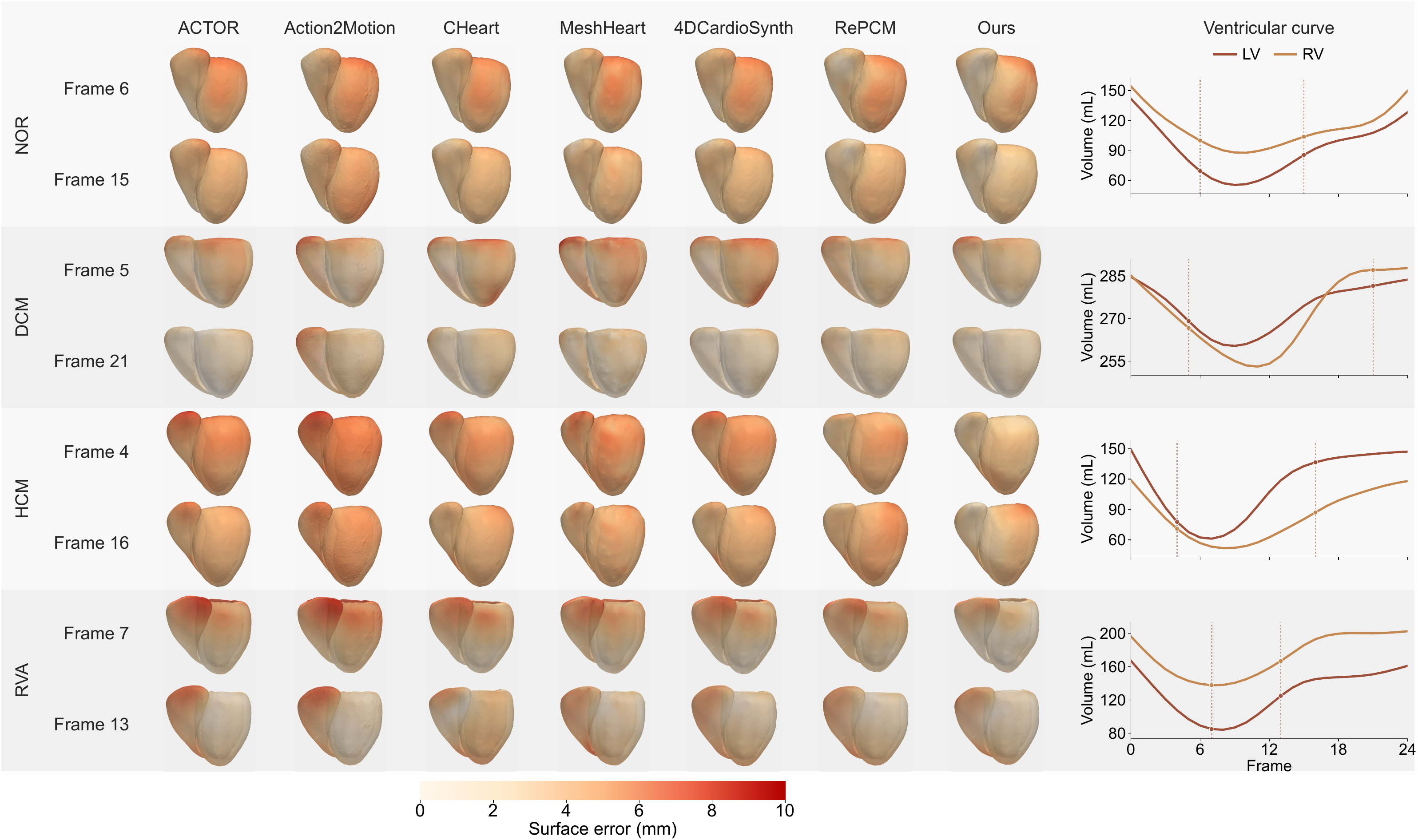}
    \caption{Qualitative comparison of ED-conditioned biventricular motion synthesis. Surface-error maps are shown for two representative cardiac phases from NOR, DCM, HCM, and RVA subjects. Each column corresponds to one competing method, and the color scale denotes the point-wise surface error relative to the reference mesh. The LV and RV volume curves on the right indicate the selected phases using vertical dashed lines.}
    \label{fig:qualitative_comparison}
\end{figure*}

\subsection{Dataset and Pre-processing}
\label{sec:data}

We used three public cine cardiac MRI datasets: ACDC~\citep{Journal/TMI/Bernard2018}, M\&Ms~\citep{Journal/TMI/Campello2021}, and M\&Ms-2~\citep{Journal/JBHI/Martin-Isla2023}. These datasets cover different acquisition protocols and cardiac conditions. 
Diagnostic labels were harmonized as normal (NOR), dilated cardiomyopathy (DCM), hypertrophic cardiomyopathy (HCM), and right ventricular abnormality (RVA). The DCM group combined ACDC/ M\&Ms DCM with M\&Ms-2 dilated-left-ventricle cases. The RVA group combined ACDC/ M\&Ms right-ventricular-abnormality cases with M\&Ms-2 arrhythmogenic-cardiomyopathy and dilated-right-ventricle cases. These groupings reflect shared LV- or RV-dominant phenotypes and were used only for stratification and subgroup evaluation.
Five additional diagnoses were reserved exclusively for diagnostic out-of-distribution (OOD) evaluation. M\&Ms hypertensive heart disease (HHD) and ischemic heart disease (IHD) were withheld because of limited sample sizes. M\&Ms-2 tetralogy of Fallot (FALL), interatrial communication (CIA), and tricuspid regurgitation (TRI) were not merged into RVA because their defining abnormalities involve anatomical structures or flow phenomena not represented by the modeled biventricular surfaces.

For each subject, the segmentation masks across the cardiac cycle were converted into a unified-topology biventricular surface-mesh sequence. We adopted an SSM-based fitting procedure based on the biventricular atlas of ~\citet{Journal/MIA/Bai2015}. The template was fitted to the multi-phase contours and refined through global alignment, non-rigid deformation, and temporal smoothing. This process produced anatomically consistent meshes with fixed topology and vertex-wise correspondence across subjects and cardiac frames. All sequences were temporally resampled to \(T=25\) frames.
For model training, all meshes were aligned to the template coordinate system through center-of-mass matching and rigid registration, followed by normalization using a global scaling factor. The same spatial transformation was applied to all frames of each subject to preserve the relative motion trajectory. During evaluation, the generated meshes were transformed back to their original physical scale before geometric and functional measurements were computed.
Following the MRI-derived phenotype definitions in~\citet{Journal/NM/Bai2020}, we extracted ED-derived phenotype descriptors from the unified biventricular meshes. The global descriptor \(c_{ED}^{global}\in\mathbb{R}^{6}\) comprised the logarithmically transformed LV ED volume, RV ED volume, and LV mass; global myocardial wall thickness; the logarithmic LV-to-RV ED volume ratio; and the logarithmic LV-mass-to-LV-volume ratio. The fine-grained descriptor \(c_{ED}^{fine}\in\mathbb{R}^{22}\) additionally included LV myocardial wall-thickness measurements from the 16 AHA regions. When functional information was available, the optional motion descriptor \(c^{motion}\in\mathbb{R}^{6}\) comprised the logarithmically transformed LV and RV end-systolic volumes, the logarithmically transformed LV and RV stroke volumes, and the LV and RV ejection fractions.
The final in-distribution cohort contained 666 subjects, as summarized in Table~\ref{exp:tb:dataset}. The data were divided at the patient level into training, validation, and test sets using a ratio of \(7{:}1{:}2\).

\subsection{Gold Standard and Evaluation}
\label{sec:evaluation}

The SSM-fitted full-cycle mesh sequences were used as the reference standard. All methods were evaluated under the same ED-conditioned synthesis setting defined in Sec.~\ref{sec:method:problem}. Before evaluation, the generated meshes were transformed back to the original physical coordinate system.
Geometric agreement between the generated and reference sequences was evaluated using the average symmetric surface distance (ASSD), the 95th-percentile Hausdorff distance (HD95), and the vertex-wise root mean square error (vRMSE). Because all meshes shared fixed vertex correspondence, vRMSE was used to quantify point-wise trajectory errors over the complete cardiac cycle.
For each test subject, every method generated 20 stochastic motion samples. Each metric was computed separately for the 20 samples and then averaged within the subject.
Functional agreement was assessed using the LV and RV volume trajectories derived from the synthesized sequences. Agreement in LVEF and RVEF was quantified using Pearson's correlation coefficient \(r\) and mean absolute error (MAE). Population-level agreement in LVESV, RVESV, LVEF, and RVEF was further evaluated using the KL divergence and Wasserstein distance.

\subsection{Implementation}
\label{sec:implementation}

All models were implemented in PyTorch and trained on a single NVIDIA RTX A5500 GPU. The motion-informed functional partition contained \(N_R=16\) regions, and the motion VAE employed \(N_a=512\) region-balanced anchor tokens. The token feature dimension and motion latent dimension were set to \(d=256\) and \(d_z=32\), respectively. Both the encoder and decoder contained \(K_m=8\) attention blocks, with four attention heads in each block. The maximum hop range and attenuation factor of MRAB were set to \(L_{\mathrm{hop}}=4\) and \(\eta=0.5\), respectively.
The motion VAE was trained for 500 epochs using Adam with a learning rate of \(1\times10^{-4}\), a batch size of 8, and a KL-divergence weight of \(\lambda_{\mathrm{KL}}=5\times10^{-4}\). 
The rectified-flow network contained \(K_r\) = 4 blocks with a hidden dimension of 128. The fine-grained descriptor \(c_{ED}^{fine}\in\mathbb{R}^{22}\) conditioned the RF blocks through AdaLN. The global descriptor \(c_{ED}^{global}\in\mathbb{R}^{6}\) was combined with the anatomical latent solely for selecting among \(N_e=4\) prototype-routed adapters. The RF model was trained for 200 epochs using AdamW with a learning rate of \(1\times10^{-5}\), a batch size of 8, and a weight decay of \(5\times10^{-3}\). During inference, the rectified-flow ODE was solved using \(N_{\mathrm{Euler}}=4\) uniform Euler steps over \(\tau\in[0,1]\).
In the optional functional-control experiments, the six-dimensional motion descriptor \(c^{motion}\) was used with a control strength of \(s=1\). This branch was disabled in the primary ED-conditioned setting.

\begin{figure*}[t]
    \centering
    \includegraphics[
        width=0.82\textwidth,
        keepaspectratio
    ]{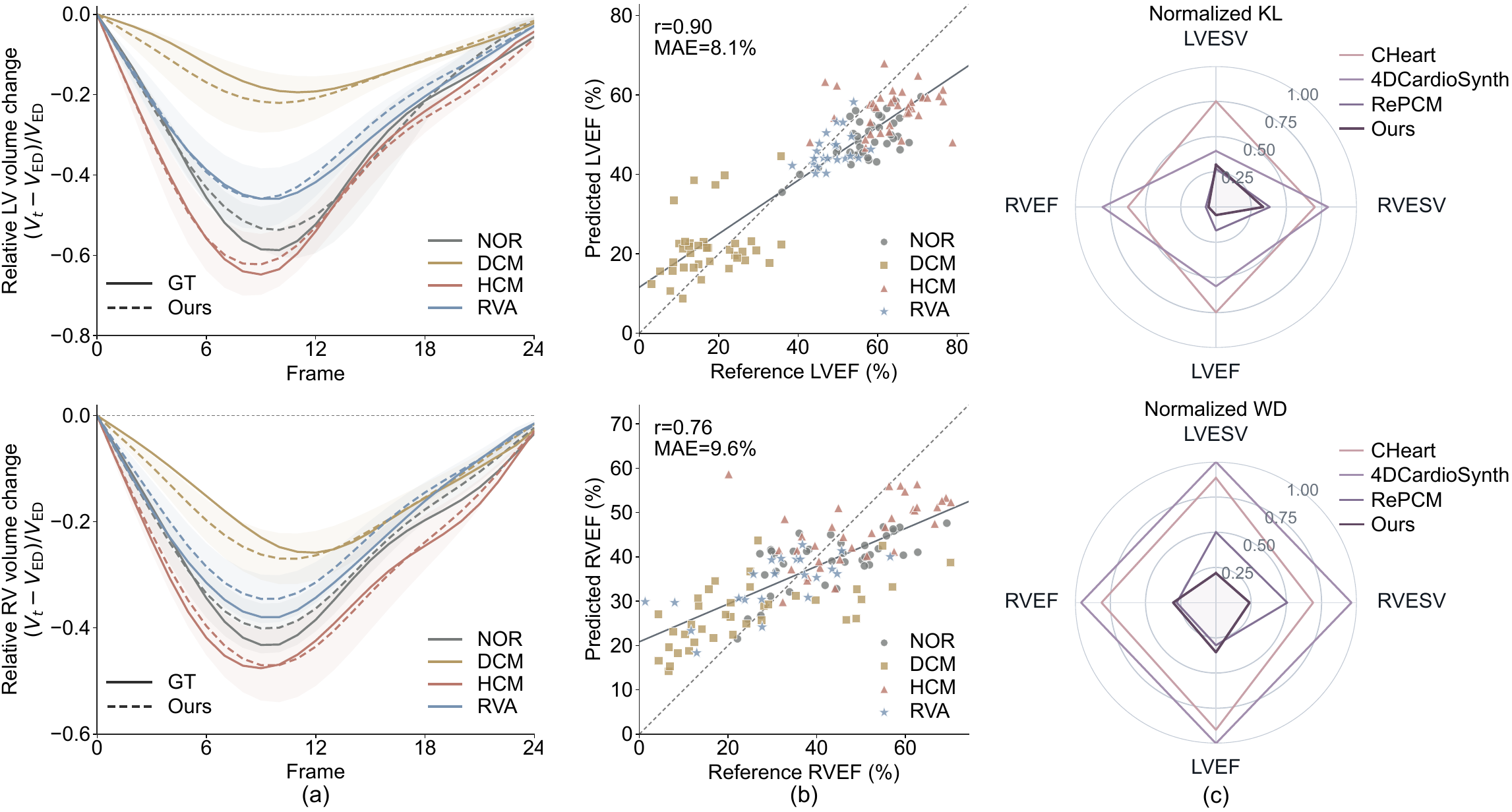}
     \caption{Functional fidelity of the synthesized biventricular sequences. (a) Diagnostic group-stratified relative volume-change trajectories for the LV (top) and RV (bottom). Solid and dashed curves represent the reference and synthesized sequences, respectively, and shaded bands indicate inter-subject variability. (b) Agreement between reference and synthesized left- and right-ventricular ejection fractions (LVEF and RVEF), with the identity line shown for reference. (c) Normalized KL divergence and Wasserstein distance (WD) between the reference and synthesized left- and right-ventricular end-systolic volumes (LVESV and RVESV), LVEF, and RVEF distributions. Lower values indicate better distributional agreement. MAE: mean absolute error.} \label{fig:functional_consistency}
\end{figure*}
\subsection{Comparison Study}  
\label{sec:main_comparison}

Table~\ref{tab:main_comparison} reports the quantitative comparison under the ED-conditioned full-cycle motion synthesis setting. 
We compared the proposed method with representative sequence generation and cardiac motion synthesis baselines, including a conditional VAE~\citep{Conference/NIPS/Sohn2015}, ACTOR~\citep{Conference/ICCV/Petrovich2021}, Action2Motion~\citep{Conference/ACMMM/Guo2020}, CHeart~\citep{Journal/TMI/Qiao2023}, MeshHeart~\citep{Journal/NMI/Qiao2025},
4DCardioSynth~\citep{Conference/MICCAI/Dou2025}, and RePCM~\citep{Conference/MICCAI/Yang2026}. For fairness, all baselines used the same pre-processed mesh sequences, ED input, temporal resolution, and patient-level split.
One can see that the proposed method achieved the best overall performance across all metrics of different chambers and their combination, indicating more accurate full-cycle motion synthesis than both generic motion-generation baselines and cardiac-specific competing methods.
Compared with all external baselines, our method consistently achieved lower biventricular (BiV), left-ventricular (LV), and right-ventricular (RV) surface and vertex-wise errors. Compared with the preliminary RePCM framework, our method significantly reduced BiV ASSD from 1.69 to 1.49 mm (\(p<0.01\)), BiV HD95 from 4.27 to 3.77 mm (\(p<0.01\)), and BiV vRMSE from 3.61 to 3.31 mm (\(p<0.05\)), as determined by subject-level paired Wilcoxon signed-rank tests. The RV results are also noteworthy because RePCM showed comparatively limited gains over the external baselines for this chamber. RV motion synthesis is particularly challenging because of the chamber’s thin myocardial wall, complex geometry, and substantial inter-subject variability. Compared with RePCM, our method reduced RV ASSD from 2.00 to 1.85 mm and RV vRMSE from 4.08 to 3.74 mm, indicating more accurate synthesis of challenging RV dynamics.

Figure~\ref{fig:qualitative_comparison} provides representative surface-error maps for all four diagnostic groups. Competing methods frequently produced spatially extended errors around the basal ventricular surfaces and the RV free wall, particularly near phases of maximal contraction. In contrast, the proposed method yielded smaller and more spatially localized errors across the selected systolic and relaxation phases. The improvement was consistent across NOR, DCM, HCM, and RVA cases, supporting the quantitative findings in Table~\ref{tab:main_comparison}.

Figure~\ref{fig:functional_consistency} presents whether the synthesized sequences preserve clinically relevant ventricular dynamics. The diagnostic group-stratified relative LV and RV volume trajectories closely followed the reference contraction-relaxation patterns, including the reduced LV contraction observed in the DCM group and the stronger relative LV volume reduction in HCM. Across individual subjects, the synthesized LVEF achieved a correlation of $r=0.90$ with an MAE of 8.1\%, while RVEF achieved $r=0.76$ with an MAE of 9.6\%. The proposed method also produced the smallest normalized KL divergence and Wasserstein distance across LVESV, RVESV, LVEF, and RVEF among the compared methods. Thus, the geometric improvements translated into more faithful ventricular volume trajectories and functional-index distributions.

\subsection{Ablation Study}  
\label{sec:ablation}
\subsubsection{Effect of Motion-Informed Parcellation}

Table~\ref{tab:region_ablation} presents the contribution of the motion-informed regional prior and the effect of partition granularity. Compared with the AHA-like partition, the motion-informed parcellation with \(N_R=8\) and \(N_R=16\) reduced all BiV errors, confirming the benefit of deriving functional regions directly from cardiac dynamics. The best overall performance was obtained with \(N_R=16\), whereas increasing the number of regions to \(N_R=32\) removed these gains. This indicates that performance does not improve monotonically with finer regionalization, as excessive fragmentation may weaken stable regional aggregation. 
\begin{figure*}[t]
    \centering
    \includegraphics[
        width=0.98\textwidth,
        keepaspectratio
    ]{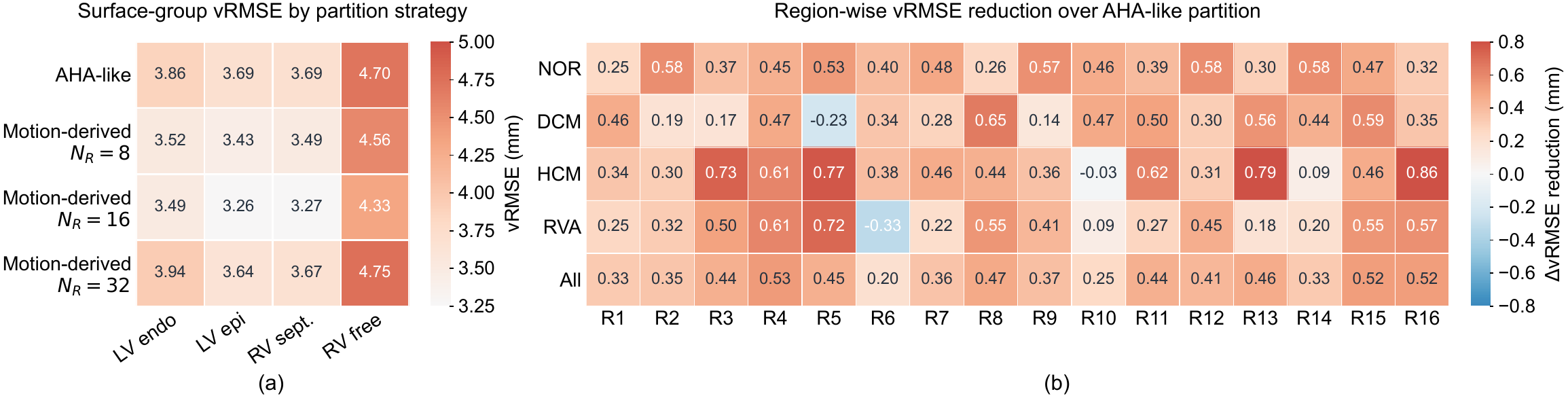}
    \caption{Regional analysis of motion-informed parcellation. (a) Surface-group vRMSE for the AHA-like partition and motion-informed parcellation with different numbers of regions. (b) Region-wise vRMSE reduction obtained by the $N_R=16$ motion-informed partition relative to the AHA-like partition, reported for each diagnostic group and the full cohort. Positive values indicate lower errors with motion-informed parcellation.} 
\label{fig:regional_analysis}
\end{figure*}

Figure~\ref{fig:regional_analysis} provides a more detailed analysis across ventricular surfaces and learned motion regions. As shown in Fig.~\ref{fig:regional_analysis} (a), the \(N_R=16\) partition achieved the lowest vRMSE across all four ventricular surface groups, while the RV free wall remained the most challenging surface under all partition strategies.
Fig.~\ref{fig:regional_analysis} (b) further reports the region-wise vRMSE reduction of the \(N_R=16\) partition relative to the AHA-like partition, where positive values indicate lower errors with motion-informed parcellation. At the cohort level, improvements were observed across all 16 learned regions. Similar improvements were found across most diagnostic group-region combinations, with only a few isolated exceptions. These results show that the benefit of motion-informed parcellation was distributed across the ventricular surface rather than being driven by a small subset of regions.

\begin{table}[t]
\centering
\caption{Effect of regional partition strategies on biventricular motion synthesis.}
\label{tab:region_ablation}
\footnotesize
\begin{tabular}{lccc}
\toprule
Partition & ASSD (mm) $\downarrow$ & HD95 (mm) $\downarrow$ & vRMSE (mm) $\downarrow$ \\
\midrule
\makecell[l]{AHA-like}
& \(1.59\pm0.44\) & \(4.11\pm1.74\) & \(3.62\pm1.43\) \\
\makecell[l]{$N_R=8$}
& \(1.50\pm0.37\) & \(3.88\pm1.39\) & \(3.40\pm1.19\) \\
\makecell[l]{\textbf{$N_R=16$}}
& \(\mathbf{1.49\pm0.34}\) & \(\mathbf{3.77\pm1.06}\) & \(\mathbf{3.31\pm1.03}\) \\
\makecell[l]{\(N_R=32\)}
& \(1.59\pm0.38\) & \(4.19\pm1.36\) & \(3.62\pm1.21\) \\
\bottomrule
\end{tabular}
\end{table}

\subsubsection{Effect of Multi-Hop Regional Attention Bias}
Table~\ref{tab:hop_ablation} summarizes the effect of the MRAB used in the motion VAE. Removing the topology bias increased BiV vRMSE from \(3.31\pm1.03\) mm to \(3.51\pm1.08\) mm, confirming the contribution of topology-guided regional communication. Introducing MRAB improved performance across most hop ranges, indicating that graded topology-aware communication among the learned regions facilitates coherent motion propagation. The \(L_{\mathrm{hop}}=4\) configuration achieved comparable mean accuracy to \(L_{\mathrm{hop}}=2\), while exhibiting consistently lower inter-subject variability across all three metrics.

\begin{table}[t]
\centering
\caption{Effect of MRAB on biventricular motion synthesis performance. MRAB: multi-hop regional attention bias.}
\label{tab:hop_ablation}

\footnotesize
\setlength{\tabcolsep}{3pt}

\resizebox{\columnwidth}{!}{%
\begin{tabular}{@{}lccc@{}}
\toprule
Topology setting 
& ASSD (mm) $\downarrow$ 
& HD95 (mm) $\downarrow$ 
& vRMSE (mm) $\downarrow$ \\
\midrule

w/o MRAB 
& \(1.56\pm0.34\) 
& \(3.92\pm1.27\) 
& \(3.51\pm1.08\) \\

\(L_{\mathrm{hop}}=1\) 
& \(1.55\pm0.42\) 
& \(3.81\pm1.71\) 
& \(3.35\pm1.38\) \\

\(L_{\mathrm{hop}}=2\) 
& \(1.49\pm0.38\) 
& \(\mathbf{3.75\pm1.55}\) 
& \(3.31\pm1.24\) \\

\(L_{\mathrm{hop}}=3\) 
& \(1.49\pm0.43\) 
& \(3.86\pm1.80\) 
& \(3.37\pm1.44\) \\

\textbf{\(L_{\mathrm{hop}}=4\)}
& \(\mathbf{1.49\pm0.34}\) 
& \(3.77\pm1.06\) 
& \(\mathbf{3.31\pm1.03}\) \\

\(L_{\mathrm{hop}}=5\) 
& \(1.55\pm0.40\) 
& \(3.94\pm1.45\) 
& \(3.41\pm1.31\) \\

\bottomrule
\end{tabular}%
}

\end{table}

\subsubsection{Effect of Phenotype Conditioning and Prototype Routing}
\label{sec:stage3_ablation}

\begin{table}[t]
\centering
\caption{Ablation of phenotype conditioning, prototype routing, and optional functional control in the rectified-flow generator. Superscripts $G$ and $F$ denote the global descriptor $c_{ED}^{global}$ and fine-grained descriptor $c_{ED}^{fine}$, respectively. The final row uses the additional motion descriptor $c^{motion}$ and is not part of the primary ED-only setting.}
\label{tab:stage3_ablation}

\footnotesize
\renewcommand{\arraystretch}{1.15}
\setlength{\tabcolsep}{2.5pt}

\resizebox{\columnwidth}{!}{%
\begin{tabular}{@{}cccccc@{}}
\toprule
$c_{ED}$ 
& Adapter 
& $c^{motion}$
& \multirow{2}{*}{ASSD (mm)$\downarrow$}
& \multirow{2}{*}{HD95 (mm)$\downarrow$}
& \multirow{2}{*}{vRMSE (mm)$\downarrow$}
\\
Condition
& Route
& Control
& & & \\
\midrule

$\times$ & $\times$ & $\times$
& $1.64\pm0.42$
& $4.17\pm1.78$
& $3.60\pm1.43$ \\

$\checkmark^{G}$ & $\times$ & $\times$
& $1.56\pm0.37$
& $4.09\pm1.53$
& $3.47\pm1.25$ \\

$\checkmark^{F}$ & $\times$ & $\times$
& $1.51\pm0.36$
& $3.92\pm1.12$
& $3.36\pm1.16$ \\

$\times$ & $\checkmark^{G}$ & $\times$
& $1.52\pm0.36$
& $3.99\pm1.47$
& $3.40\pm1.16$ \\

$\times$ & $\checkmark^{F}$ & $\times$
& $1.54\pm0.37$
& $3.97\pm1.44$
& $3.43\pm1.17$ \\

\midrule

$\checkmark^{G}$ & $\checkmark^{G}$ & $\times$
& $1.51\pm0.36$
& $3.83\pm1.52$
& $3.41\pm1.19$ \\

$\checkmark^{G}$ & $\checkmark^{F}$ & $\times$
& $1.56\pm0.35$
& $4.00\pm1.23$
& $3.54\pm1.18$ \\

$\checkmark^{F}$ & $\checkmark^{G}$ & $\times$
& $\mathbf{1.49\pm0.34}$
& $\mathbf{3.77\pm1.06}$
& $\mathbf{3.31\pm1.03}$ \\

$\checkmark^{F}$ & $\checkmark^{F}$ & $\times$
& $1.54\pm0.37$
& $3.97\pm1.44$
& $3.39\pm1.16$ \\

\midrule

$\checkmark^{F}$ & $\checkmark^{G}$ & $\checkmark$
& $1.46\pm0.39$
& $3.74\pm1.49$
& $3.21\pm1.23$ \\

\bottomrule
\end{tabular}%
}

\end{table}

\begin{figure}[t]
    \centering
    \includegraphics[
        width=0.96\columnwidth,
        keepaspectratio
    ]{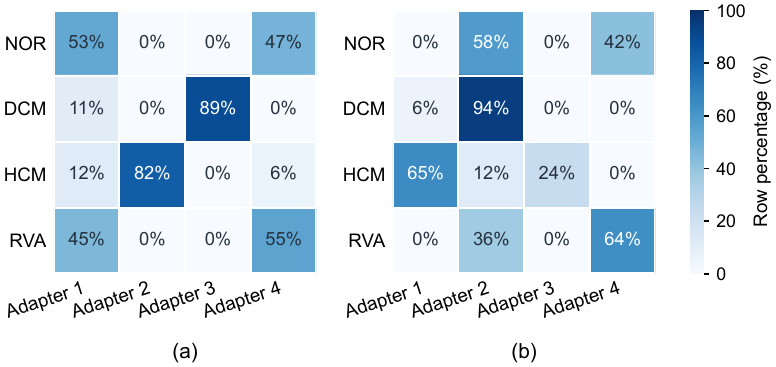}
     \caption{Diagnostic-group distributions across prototype-routed motion adapters. Row-normalized adapter assignments are shown for routing using the anatomical latent together with (a) the global ED descriptor \(c_{ED}^{global}\) or (b) the fine-grained ED descriptor \(c_{ED}^{fine}\). Each cell reports the percentage of subjects in a diagnostic group assigned to the corresponding adapter.} 
    \label{fig:routing_analysis}
\end{figure}

Table~\ref{tab:stage3_ablation} evaluates the two descriptor-based conditioning mechanisms used in Sec.~\ref{sec:method:motion_gen}. Relative to the baseline without explicit phenotype conditioning or routing, fine-grained ED conditioning consistently improved all three metrics, while global-descriptor-based routing also yielded clear gains across the three measures. Combining $c_{ED}^{fine}$ for latent conditioning with $c_{ED}^{global}$ for adapter routing achieved the best ED-only result, with an ASSD of $1.49$~mm, HD95 of $3.77$~mm, and vRMSE of $3.31$~mm. The two mechanisms therefore provide complementary benefits: fine-grained descriptors guide subject-specific latent generation, whereas global descriptors provide a more discriminative signal for routing among the motion adapters.

Figure~\ref{fig:routing_analysis} explains the different roles of the two descriptors. Routing based on $c_{ED}^{global}$ assigned DCM and HCM subjects predominantly to distinct adapters, while NOR and RVA subjects were distributed mainly across the remaining routes. In contrast, routing with $c_{ED}^{fine}$ produced greater overlap across diagnostic groups and distributed HCM subjects across several adapters. Fine-grained descriptors are therefore useful as continuous generation conditions, but may contain route-irrelevant local variation that weakens the separation of global motion modes. Accordingly, the final model uses $c_{ED}^{fine}$ for latent conditioning and $c_{ED}^{global}$ for prototype routing.

\begin{table}[t]
\centering
\caption{Effect of the number of routing prototypes/ adapters \(N_e\) on biventricular motion synthesis performance. }
\label{tab:adapter_number}
\footnotesize
\begin{tabular}{cccc}
\toprule
\(N_e\) & ASSD (mm) $\downarrow$ & HD95 (mm) $\downarrow$ & vRMSE (mm) $\downarrow$ \\
\midrule
\(\mathbf{4}\) & \(1.49\pm0.34\) & \(\mathbf{3.77\pm1.06}\) & \(\mathbf{3.31\pm1.03}\) \\
6 & \(\mathbf{1.47\pm0.38}\) & \(3.82\pm1.58\) & \(3.36\pm1.30\) \\
8 & \(1.50\pm0.40\) & \(3.94\pm1.72\) & \(3.41\pm1.38\) \\
\bottomrule
\end{tabular}
\end{table}

\begin{figure}[t]
    \centering
    \includegraphics[
        width=\columnwidth,
        keepaspectratio
    ]{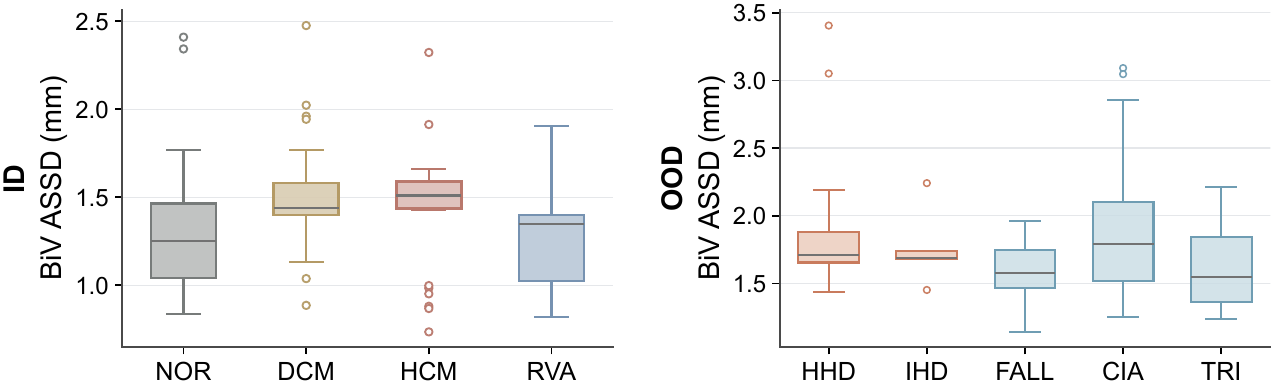}
    \caption{Geometric errors across in-distribution (ID) and out-of-distribution (OOD) diagnostic groups. (a) Four ID groups shared by ACDC, M\&Ms, and M\&Ms-2. (b) Five OOD groups excluded from training: hypertensive heart disease (HHD) and ischemic heart disease (IHD) from M\&Ms, and tetralogy of Fallot (FALL), interatrial communication (CIA), and tricuspid regurgitation (TRI) from M\&Ms-2.} 
    \label{fig:diseasewise_ood}
\end{figure}

\begin{figure}[h!]
    \centering
    \includegraphics[
        width=0.9\columnwidth,
        keepaspectratio
    ]{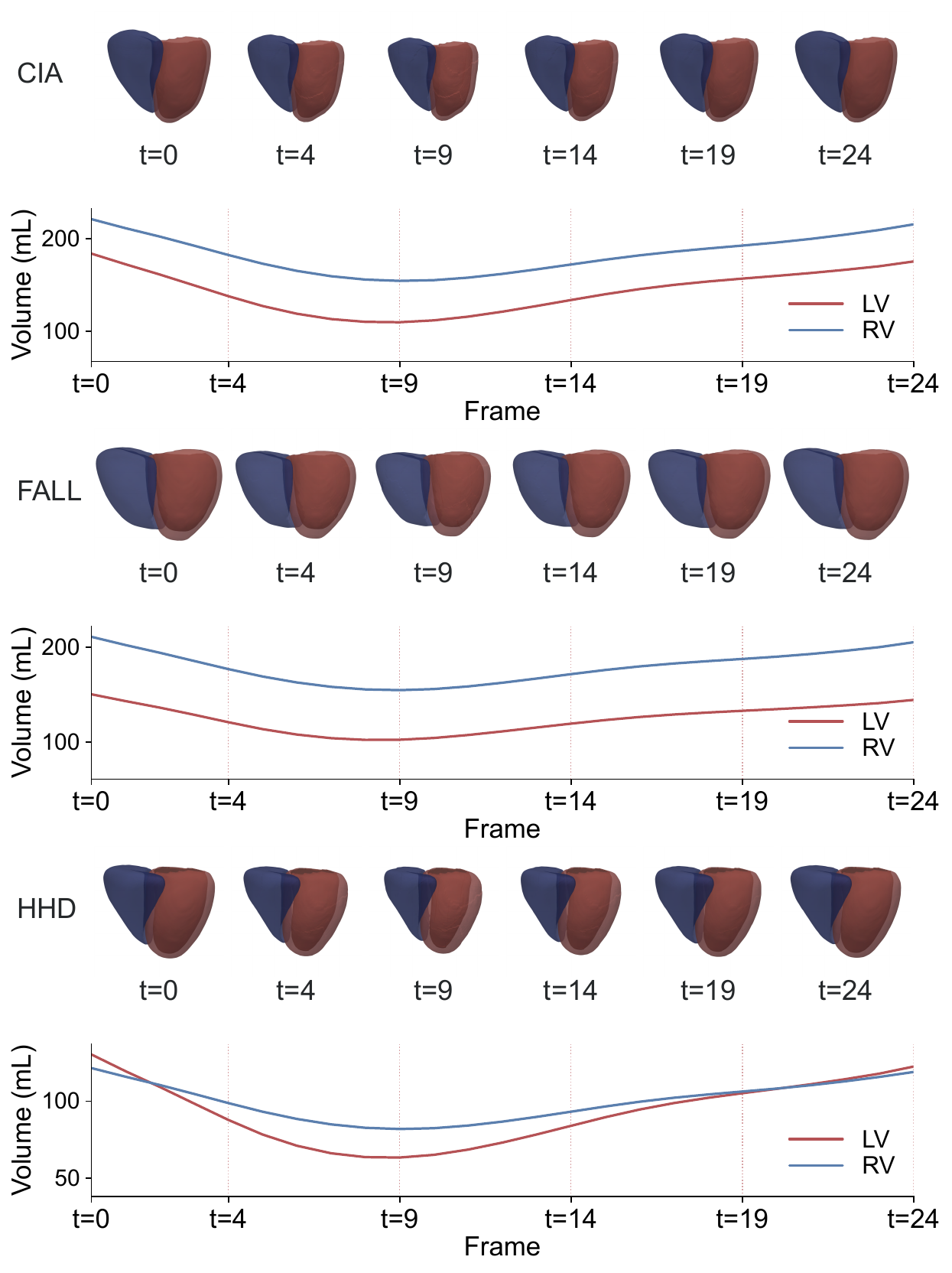}
    \caption{Illustration of full-cycle synthesized mesh sequences for three representative OOD cases and their corresponding LV and RV volume trajectories.} 
    \label{fig:external_validation}
\end{figure}

\begin{figure}[h!]
    \centering
    \includegraphics[
        width=\columnwidth,
        keepaspectratio
    ]{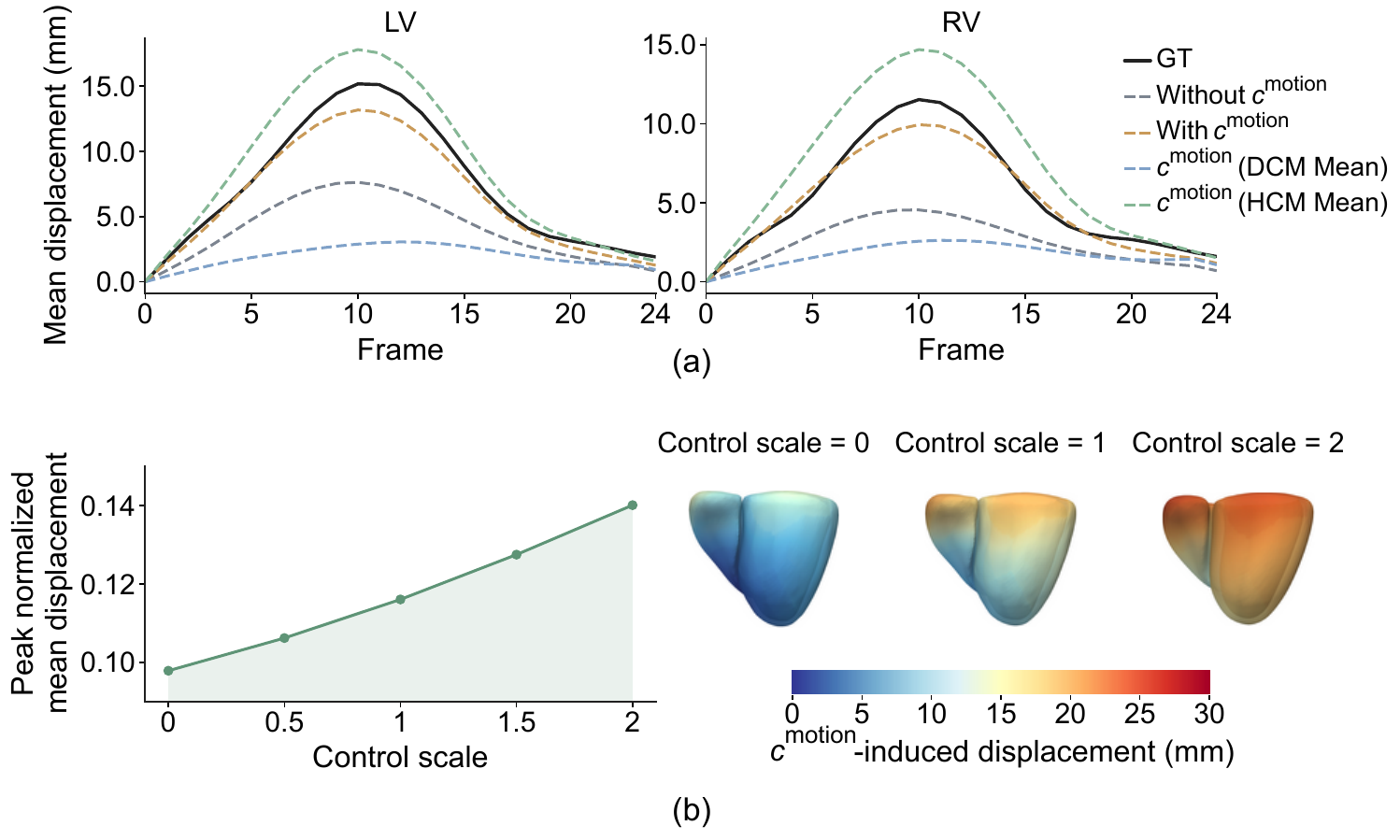}
    \caption{Controllable modulation of synthesized cardiac motion. (a) Mean LV and RV displacement trajectories for the reference, the ED-only synthesis, and predictions obtained using the optional functional-control branch. With \(c^{motion}\) uses the subject-specific motion descriptor of the illustrated NOR case, whereas the DCM-mean and HCM-mean curves use the corresponding group-average descriptors. (b) Response of a representative NOR case to control scale ranging from 0 to 2, including the relationship between the control coefficient and peak normalized mean displacement and the corresponding spatial maps of the \(c^{motion}\)-induced deformation at \(s=0\), \(s=1\), and \(s=2\).} 
    \label{fig:motion_control}
\end{figure}


Table~\ref{tab:adapter_number} evaluates the sensitivity to the number of routing prototypes/ adapters \(N_e\). Performance was relatively stable across different values of \(N_e\). Although \(N_e=6\) achieved a slightly lower ASSD, \(N_e=4\) produced the best HD95 and vRMSE with lower variance. We used \(N_e=4\) as the default setting, providing a compact routing module while maintaining stable motion synthesis accuracy.

\subsection{In- and Out-of-Distribution Diagnostic-Group Analysis}

Figure~\ref{fig:diseasewise_ood} (a) reports the geometric errors across four in-distribution (ID) diagnostic groups shared by ACDC, M\&Ms, and M\&Ms-2. Performance remained broadly stable across the four groups, with no marked degradation observed in any single group. DCM and HCM exhibited somewhat higher median ASSD than NOR and RVA, reflecting the greater variability associated with pathological ventricular remodeling and contraction. Nevertheless, their interquartile ranges remained substantially overlapped.
Fig.~\ref{fig:diseasewise_ood} (b) further evaluates five OOD diagnostic groups from M\&Ms and M\&Ms-2 that were completely excluded from training: HHD, IHD, FALL, CIA, and TRI. Although some OOD groups exhibited greater error variability, particularly CIA, the overall error distributions remained within a comparable range across the two datasets. These results suggest that the learned motion representation generalizes beyond the diagnostic groups observed during training rather than relying on a single dominant disease pattern.
Representative synthesized sequences are shown in Fig.~\ref{fig:external_validation} for OOD cases with CIA, FALL, and HHD. The generated biventricular surfaces evolve smoothly throughout the cardiac cycle and exhibit distinct LV and RV volume trajectories across the three cases. These examples illustrate that the model generates temporally coherent motion with distinct case-specific ventricular dynamics across the three OOD cases.

\subsection{Optional Functional Motion Control}
\label{sec:motion_control}

The primary model infers motion solely from ED anatomy and ED-derived phenotype descriptors. When additional motion descriptors are available, the optional \(c^{motion}\) branch enables explicit modulation of the generated motion. As shown in the final row of Table~\ref{tab:stage3_ablation}, incorporating this control reduced BiV ASSD from \(1.49\) to \(1.46\)~mm and vRMSE from \(3.31\) to \(3.21\)~mm.
Fig.~\ref{fig:motion_control} (a) shows that the controlled LV and RV mean-displacement trajectories more closely followed the reference than the uncontrolled synthesis. Conditioning on the mean motion descriptors derived from the DCM and HCM groups produced distinct displacement amplitudes, demonstrating descriptor-dependent modulation of cardiac motion. Fig.~\ref{fig:motion_control} (b) further evaluates a representative NOR case under control scale ranging from 0 to 2. The peak normalized mean displacement increased monotonically and approximately linearly with the control scale, while the corresponding spatial maps showed smooth and spatially varying deformation. These results indicate that the optional branch provides interpretable control over motion magnitude without changing the primary ED-only inference setting.

\section{Discussion and Conclusion}

This study presents a region-specific and phenotype-adaptive framework for synthesizing full-cycle biventricular motion from a single ED mesh. Across three public cine MRI datasets and four in-distribution diagnostic groups, the proposed ED-conditioned model consistently outperformed generic motion generators and cardiac-specific synthesis methods in surface- and correspondence-based metrics. The improvement over RePCM indicates that replacing global latent generation with structured motion representation and phenotype-conditioned RF generation provides a more effective formulation for this task. The gains were particularly evident for the RV, whose complex geometry and greater inter-subject variability make its motion more difficult to recover. Importantly, the geometric improvements were accompanied by better functional fidelity: the synthesized ventricular volume trajectories preserved diagnostic-group-specific ventricular motion patterns, and the derived LVEF and RVEF showed meaningful agreement with their reference values. The lower distributional discrepancies in ventricular volumes and ejection fractions further suggest that the model captures clinically relevant population-level variation rather than merely minimizing point-wise geometric errors. The OOD evaluation on M\&Ms and M\&Ms-2 further indicates that the learned representation generalizes beyond the diagnostic groups observed during training, although the increased variability in some OOD groups highlights the remaining challenge of uncommon pathological anatomies.

The ablation studies provide complementary evidence for the principal design choices. Motion-informed functional parcellation consistently outperformed the AHA-like partition, demonstrating that regions learned directly from deformation patterns are better suited to generative motion modeling than predefined anatomical divisions. The strongest performance was obtained with 16 regions, whereas excessive partitioning weakened regional aggregation, indicating that the regional prior should balance local specificity with motion coherence. MRAB further improved synthesis by permitting graded communication between neighboring functional regions while suppressing unrelated interactions. In the latent generator, fine-grained ED phenotype descriptors were more effective for conditioning motion generation, whereas global ED phenotype descriptors produced clearer and more stable prototype routing. This distinction suggests that detailed regional morphology is useful for specifying subject-level dynamics, while global anatomy is better suited to identifying broader motion modes. The optional functional-control branch further modulated motion magnitude in a continuous manner, but its results should be interpreted separately from the primary ED-only setting because it requires additional motion descriptors unavailable from the ED anatomy alone.

Several limitations remain. First, the reference sequences were obtained through SSM fitting of segmentation masks and therefore inherit errors introduced by image segmentation, template fitting, and temporal smoothing; evaluation against independently tracked or manually verified 4D meshes would provide a stronger assessment of motion accuracy. Second, the current study focuses on unified-topology biventricular surfaces and four relatively broad diagnostic groups. Its applicability to atrial motion, myocardial-layer deformation, congenital abnormalities, and more heterogeneous or subtle disease subtypes remains to be investigated. Third, although ED anatomy provides an accessible patient-specific condition, it cannot uniquely determine functional properties such as contraction strength, electromechanical delay, or regional dysfunction, as also reflected by the remaining errors in RVEF estimation. Future work should therefore incorporate complementary imaging, clinical, or physiological conditions when available, evaluate generalization on independent clinical cohorts and alternative mesh-construction pipelines, and extend the framework toward controllable whole-heart motion generation and virtual population synthesis for downstream computational modeling and in silico studies.

\bibliographystyle{model2-names}
\biboptions{authoryear}
\bibliography{A_refs}

\end{document}